\documentclass{article}
\usepackage{spconf,amsmath,graphicx}
\usepackage{booktabs}
\usepackage{hyperref}
\usepackage{multirow}
\usepackage{diagbox}
\usepackage[normalem]{ulem}
\useunder{\uline}{\ul}{}
\usepackage{footnote}

\usepackage{svg}
\usepackage{enumitem}

\title{Weighted Sampling For Masked Language Modeling}
\name{Linhan Zhang$^\dag$, Qian Chen$^\ddag$, Wen Wang$^\ddag$, Chong Deng$^\ddag$, Xin Cao$^\dag$, Kongzhang Hao$^\dag$, Yuxin Jiang$^*$, Wei Wang$^*$}

\address{$\dag$ University of New South Wales, School of Computer and Engineering\\
        $\ddag$ Speech Lab of DAMO Academy, Alibaba Group \\
        $*$ Hong Kong University of Science and Technology (Guangzhou), China \\
            \tt \normalsize \{linhan.zhang,xin.cao\}@unsw.edu.au \\
            \tt \normalsize \{tanqing.cq,w.wang,dengchong.d\}@alibaba-inc.com \\ 
            \tt \normalsize weiwcs@ust.hk}
\begin{document}
\ninept
\maketitle
\begin{abstract}
% Masked Language Modeling (MLM) is widely used for pre-training language models. The standard random masking strategy in MLM causes the pre-trained language models (PLMs) biased toward high-frequency tokens. Representations of rare tokens are poorly learned and the performance of PLMs on downstream tasks is limited.
% To alleviate this frequency bias issue, we propose two simple and effective \textbf{Weighted Sampling} strategies for masking tokens based on token frequency and training loss. We apply these two strategies to BERT and the resulting pre-trained model is denoted by \textbf{Weighted-Sampled BERT (WSBERT)}.  Experiments on the Semantic Textual Similarity benchmark (STS) show significant improvement on sentence embeddings from WSBERT over BERT. Combining WSBERT with other calibration methods and prompt learning further improves sentence embeddings. We also investigate fine-tuning WSBERT on the GLUE benchmark and show that Weighted Sampling also improves the transfer learning capability of the backbone PLM. Meanwhile, we conduct analysis and provide insights on how WSBERT improves the token embeddings.
Masked Language Modeling (MLM) is widely used to pretrain language models. The standard random masking strategy in MLM causes the pre-trained language models (PLMs) to be biased towards high-frequency tokens. Representation learning of rare tokens is poor and PLMs have limited performance on downstream tasks.
To alleviate this frequency bias issue, we propose two simple and effective \textbf{Weighted Sampling} strategies for masking tokens based on token frequency and training loss. We apply these two strategies to BERT and obtain \textbf{Weighted-Sampled BERT (WSBERT)}. Experiments on the Semantic Textual Similarity benchmark (STS) show that WSBERT significantly improves sentence embeddings over BERT. Combining WSBERT with calibration methods and prompt learning further improves sentence embeddings. We also investigate fine-tuning WSBERT on the GLUE benchmark and show that Weighted Sampling also improves the transfer learning capability of the backbone PLM. We further analyze and provide insights into how WSBERT improves token embeddings.
%Conventional pre-trained language models trained with masking strategies are generally biased towards masking high-frequency tokens. This causes insufficient learning on rare tokens and leads to poorer performance on downstream tasks. To alleviate the frequency bias issue, we propose two simple and effective \textbf{weighted sampling} strategies for sampling masked tokens during continual training of any pre-trained language model trained with masking strategies. We apply these two strategies on BERT and denote the resulting pre-trained model by \textbf{Weighted-Sampled BERT (WSBERT)}.  Experiments on the Sentence Similarity benchmark (STS) show the effectiveness of WSBERT as an unsupervised sentence representation model. Combining WSBERT with other calibration approaches and prompt learning further improves sentence representation modeling. We also investigate fine-tuning WSBERT on GLUE and show that the weighted sampling strategies could enhance the transfer learning capability of the backbone language model.
%and propose a model denoted as Weighted-Sampled BERT (WSBERT). To evaluate efficiency  of WSBERT, we explore the performance of WSBERT on the Sentence Similarity benchmark (STS). Besides, we also propose to aggregate some augmentation methods compatible with WSBERT to further improve its performance on STS. Furthermore, we investigate the performance of  WSBERT on GLUE benchmarks, which proves weighted sampling could enhance its backbone language model. 
\end{abstract}
\vspace{-3mm}
\section{Introduction}
\label{sec:intro}
\vspace{-2mm}
Early language models model context unidirectionally, either left-to-right or right-to-left. In contrast, Masked Language Modeling (MLM) replaces a subset of tokens in the input sequence with a special token \verb![MASK]! and trains the model to predict the masked tokens using their bidirectional context. MLM has been widely adopted as a self-supervised pre-training objective for learning bidirectionally contextualized language representations, such as BERT~\cite{bert} and RoBERTa~\cite{roberta}. BERT and its extensions as pre-trained language models (PLMs) have shown remarkable performance on various downstream NLP tasks. 

\begin{figure}[t]
    \centering
    \includegraphics[width=0.6\linewidth]{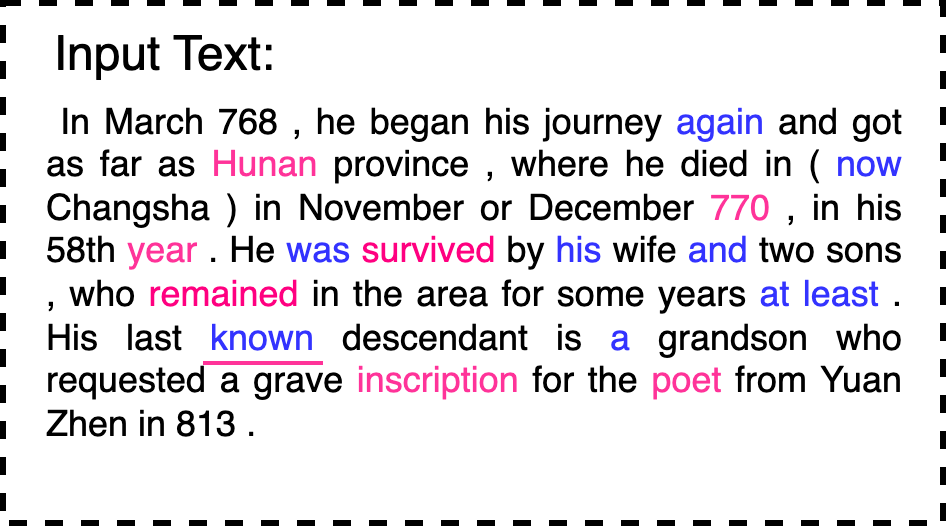}
    \caption{\small{An example from WikiText. Randomly selected tokens are in blue while Frequency Weighted Sampled tokens are in pink.}}
    \label{fig:input}
\end{figure}

Nevertheless, recent studies reveal critical problems in MLM. \cite{representation_degeneration, how_contextual} find the contextualized word representations of BERT and other PLMs are not isotropic as they are not uniformly distributed w.r.t. direction; instead, they are anisotropic as word representations occupy a narrow cone. The token frequency in the pre-training data usually follows a long-tailed distribution. The conventional masking strategy for MLM selects tokens to mask with a uniform distribution~\cite{bert,roberta}. 
This random masking strategy for MLM unavoidably encounters the frequency bias issue, that is, high-frequency tokens will be masked frequently, while more informative tokens, typically with lower frequencies, will be masked much less frequently during pre-training, which would greatly harm the efficiency of pre-training, lower the quality of representations of rare tokens and limit the performance of PLMs. As shown in Figure~\ref{fig:input}, tokens selected based on their frequency (in pink, see Eqn.~\ref{eq:freq_weight}) are apparently more informative than tokens selected randomly (in blue) which are mostly high-frequency tokens but not essential to the semantics of the sentence. \cite{on_the_sentence} investigates the embedding space of MLM-trained PLMs and confirms that embeddings are biased by token frequency and rare tokens are distributed sparsely in the embedding space. \cite{promptbert} demonstrates that frequency bias indeed harms the performance of sentence embeddings generated by MLM-trained PLMs.
As shown in these studies, alleviating the frequency bias issue is essential for improving effectiveness of MLM and performance of resulting PLMs.

Several recent studies focus on improving efficiency of pre-training, including mixed-precision training~\cite{megatron-lm}, parameter distillation for different layers~\cite{effecient_training}, introducing a note dictionary for saving information of rare tokens~\cite{tfn}, designing different training objectives~\cite{albert,electra,cocolm}, and dropping redundant tokens during pre-training~\cite{token_dropping}. However, most of these approaches focus on modifying model architecture or optimization for pre-training. 

Our work focuses on alleviating the frequency bias issue in MLM and improving quality of PLMs. We propose two \textbf{Weighted Sampling} methods for masking tokens based on token frequency or training loss. The latter one can dynamically adjust sampling weights and achieve a good balance between masking probabilities of \emph{common tokens} and \emph{rare tokens}\footnote{We denote high-frequency and low-frequency tokens by \emph{common tokens} and \emph{rare tokens} in the rest of the paper.} based on the learning status of PLMs. 
Our Weighted Sampling methods can be applied to \emph{any} MLM-pretrained PLMs. In this work, we focus on investigating the effectiveness of applying Weighted Sampling to BERT as the backbone. We initialize from BERT and continue pre-training with Weighted Sampling. We denote the resulting PLM by \textbf{WSBERT}. We hypothesize that since Weighted Sampling could alleviate frequency bias, it could improve representation learning of rare tokens and also improve the overall quality of language representations. Quality of pre-trained language representations is generally evaluated on sentence representations generated by PLMs, commonly evaluated on the Semantic Textual Similarity (STS) benchmark \cite{semeval2012,semeval2013,semeval2014,semeval2015,semeval2016,semeval2017,sick}; and evaluated on transfer learning capability of PLMs, commonly evaluated on fine-tuning and testing on the GLUE benchmark~\cite{glue}.  Recent efforts on sentence representation modeling include calibration methods~\cite{on_the_sentence,whitening}, prompt learning\cite{radford2018gpt,radford2019gpt,lm_few_shot,cloze_question, making_plm_few_shot, promptbert},  and sentence-level contrastive learning (CL) based models such as SimCSE~\cite{simcse} and its variants~\cite{continuous_prompt,infocse}. Although SimCSE and its variants achieve state-of-the-art (SOTA) performance on STS, 
they degrade the transfer learning capability on tasks such as SQuAD since they do not target improving token-level representation learning~\cite{tacl}. We also observe absolute 0.5 performance degradation on GLUE from SimCSE-BERT compared to BERT. 

In this work, to investigate whether the proposed Weighted Sampling could improve the quality of token embeddings, we evaluate sentence representations generated by WSBERT on STS and the transferability of WSBERT on GLUE. We also analyze the embedding space of WSBERT and BERT to understand how Weighted Sampling improves the quality of token embeddings.
Our contributions can be summarized as follows:
\begin{itemize}[leftmargin=*,noitemsep]
    \item We propose two \textbf{Weighted Sampling} methods to alleviate the frequency bias issue in conventional masked language modeling.
    \item We develop a new PLM, WSBERT, by applying Weighted Sampling to BERT. Different from SOTA sentence representation models, we find WSBERT outperforms BERT on \textbf{both sentence representation quality and transfer learning capability}. We also find integrating calibration methods and prompts into WSBERT further improve sentence representations.
    \item We design ablation approaches to analyze the embedding space of WSBERT and BERT. We find that with Weighted Sampling, rare tokens are more concentrated with common tokens and common tokens are more concentrated in the embedding space than BERT. We also find that both common and rare tokens are closer to the origin in WSBERT than BERT and token embeddings of WSBERT are less sparse than BERT. We believe these improvements in token embeddings caused by Weighted Sampling contribute to the improvements in sentence representations and transferability.
\end{itemize}

\vspace{-3mm}
\section{Method}
\label{sec:method}
\vspace{-2mm}
In this section, we first describe traditional Masked Language Modeling (MLM). Then we propose \textbf{Weighted Sampling} for MLM to alleviate the frequency bias problem.
% Although pretrained MLM has achieved promising performance in most of NLP-related tasks, there is a frequency bias among most MLM not addressed, which has weakened their performance. 
%%%%In this section, we first introduce the mask language model (MLM), and then we propose a training strategy named weighted sampling for MLM to alleviate the influence of frequency bias. 
% Additionally, we also develop a novel prompt-based method to improve the quality of sentence representation generated from weighted-sampled MLM by smoothing the distribution of sentence embedding generated by MLM. 
%%% 这段需要重新写
\vspace{-3mm}
\subsection{Masked Language Modeling}
%\subsection{Mask Language Model} %
%%!!  not ...
For a sentence $S=\{t_1, t_2, \ldots, t_n\}$, where $n$ is the number of tokens and $t_i$ is a token, the standard masking strategy as in~\cite{bert} randomly chooses 15\% of tokens to mask.  The language model learns to predict the masked tokens with bidirectional context. To make the model compatible with fine-tuning, for a chosen token, 10\% of the time it is replaced by a random token from the corpus, 10\% of the time it remains unchanged, and 80\% of the time it is replaced by a special token \verb![MASK]!.
% {\sf The probability of each token to be chosen is named masking probability in this paper. }%%!! awkward
% The language model only predicts the masked tokens instead of reconstructing the whole input sequence. %

%%%%The language model learns to predict the masked tokens. To make the model compatible with the fine-tuning task, 10\% of the above-chosen tokens will be masked by some actual tokens token from the corpus, 10\% tokens remain unchanged, and the remaining 80\% will be replaced by a special token \verb![MASK]!.

\subsection{Weighted Sampling}
%%%%\subsection{BERT with Weight Sampling}
In order to tackle the frequency bias problem, we propose two weighted sampling strategies, namely, \textbf{Frequency Weighted Sampling} and \textbf{Dynamic Weighted Sampling}, to compute the masking probability for each 
%unique
token, based on statistical signals and model-based signals, respectively.

%%%%Recall that MLM randomly masks tokens, there is an inherent bias that frequently occurring tokens will be masked more often than rare tokens; this will greatly lower the difficulty of the pretraining task and may make the model stop prematurely. As a consequence, stopwords such as ``at'' or ``and'' in Figure \ref{fig:input} will be masked frequently, while more informative tokens, typically with lower frequencies, may never be masked during the learning. 
%%!! too late. need to say this earlier.
%The frequency bias will greatly lower the difficulty of the pretraining task. %
%In this way, the language model could not be fully trained and optimized. 

%%%%To deal with the frequency bias, we propose to adopt a weighted sampling strategy for the masking process. %
% {\sf In the beginning, we use a weight dictionary whose keys are tokens in training vocabulary and the values are masking probability.}\fixme{not important at this stage; do not distract the reader}
%
%%%%We design two strategies to calculate the masking probability for each unique token, based on statistical signals and model-specific signals, respectively. % frequency masking and dynamic masking. 
\vspace{-2mm}
\subsubsection{Frequency Weighted Sampling}
%%%%\subsubsection{Frequency Masking}
A natural statistical signal characterizing the informativeness of a token $w$ is its frequency $\mathrm{freq}(w)$ in the pre-training corpus. %
We first apply the following transformation to remove the excessive influences of extremely rare tokens, which are usually noise.
\begin{equation}
    \mathrm{freq}^*(w) = 
    \begin{cases}
        \mathrm{freq}(w) & \text{, if } \mathrm{freq}(w) > \theta \\
        \theta & \text{, otherwise.}
    \end{cases}
\end{equation}
\noindent Then we compute the sampling weight $wt(w)$ for $w$ as follows. 
\begin{equation}
    \label{eq:freq_weight}
    \mathrm{wt}(w) = (\mathrm{freq}^*(w))^{-\alpha}
\end{equation}
\noindent In our experiments, we set the hyperparameters $\theta=10$ and $\alpha=0.5$ based on optimizing performance on the development set.

%%%%We first apply the following transformation to remove the excessive influences of extremely rare tokens, which are usually noise.

%%%%A natural statistical signal that characterizes the informativeness of a token $w$ is its frequencies $f_w$. % To obtain the sampling probabilities for each token, we first apply the following  transformation to remove the excessive influences of extremely rare tokens. 
%%%%\begin{equation}
%%%%    \mathrm{freq}^*(w) = 
%%%%    \begin{cases}
%%%%        \mathrm{freq}(w) & \text{, if } \mathrm{freq}(w) > \theta \\
%%%%        \theta & \text{, otherwise.}
%%%%    \end{cases}
%%%%\end{equation}

%%%%Then, we obtain the weight for $w$ as $wt(w) = (\mathrm{freq}^*(w))^{-\alpha}$. In our experiments, we found that $\theta = 10$ and $\alpha = 0.5$ works well. 
For each token $t_i$ in a sentence $S=\{t_1, t_2, \ldots, t_n\}$, where $n$ is the number of tokens, we compute the sampling probability $p(t_i)$ for masking $t_i$ by normalizing $wt(t_i)$.
\begin{equation}
\label{eq:freq_prob}
\small
    p(t_i) = \frac{wt(t_i)}{\sum_{j=1}^{n}{wt(t_j)}}
\end{equation}
%\begin{equation}
%\label{eq:freq}
%\small
%    p(t_i) = \frac{\sum_{t_i = w} wt(w)}{\sum_{j=1}^{n}{wt(t_j)}}
%\end{equation}

\begin{figure}[]
  \centering
  \centerline{\includegraphics[width=0.6\linewidth]{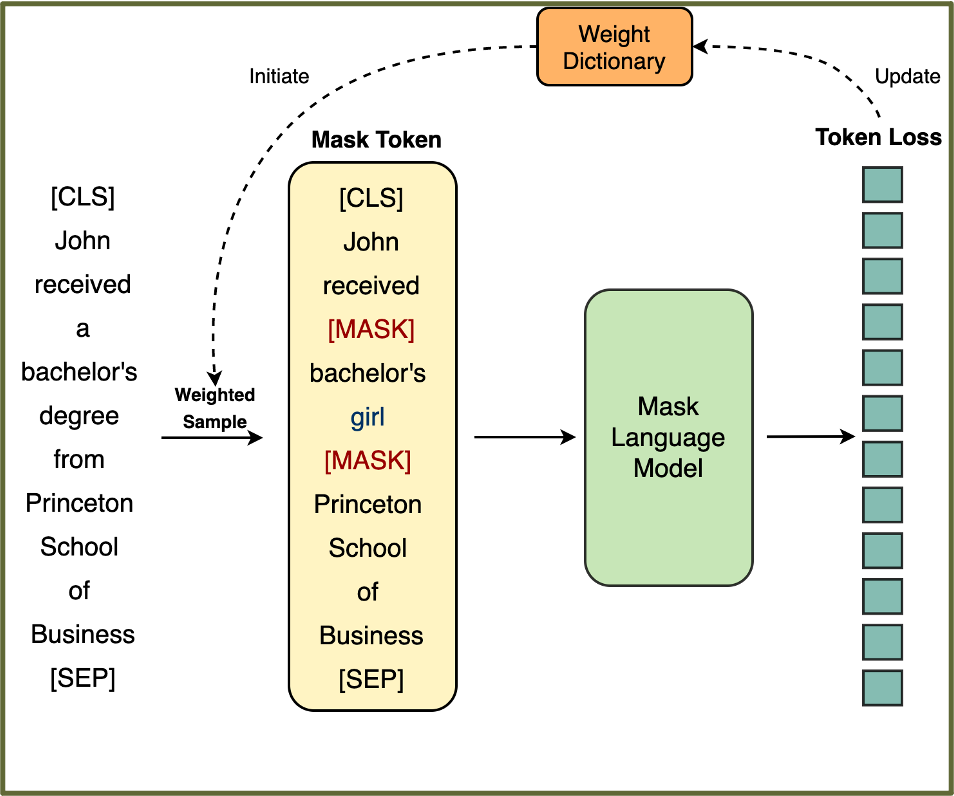}}
%  \vspace{2.0cm}
\caption{\small{Illustration of the proposed \textbf{Dynamic Weighted Sampling} for mask language modeling (MLM). The sampling weight of choosing a token to mask is computed based on the prediction loss of this token by the current PLM. We store the sampling weights of each token in the weight dictionary.}}
\label{fig:example}
\end{figure}

\vspace{-2mm}
\subsubsection{Dynamic Weighted Sampling}
%%%%\subsubsection{Dynamic Masking}
Frequency Weighted Sampling produces constant sampling probabilities for tokens and does not consider the learning status of the backbone masked language model that it applies for.  We hypothesize that the signal of informativeness of a token $w$ may also be derived from how poorly a masked language model predicts it. Therefore, we propose the Dynamic Weighted Sampling strategy shown in Figure~\ref{fig:example}. We use a weight dictionary in memory to store the sampling weights of each token after each batch in each iteration instead of updating the sampling weights after processing all batches in an iteration as the sampling only happens once in the latter case. Firstly, we set an initial sampling weight
% We use a weight dictionary in memory to store the sampling weights of each token after each batch in each iteration instead of processing all batches in an iteration for an update, in which the sampling only happens once.
%masking probability
$wt(t_i)=1$ for each token $t_i \in T$ in the weight dictionary, where $T$ denotes all tokens in the pre-training dataset. Then, we compute the sampling probabilities using sampling weights in the weight dictionary based on Eqn.~\ref{eq:freq_prob} and train a masked language model. During each mini-batch, the masked tokens are predicted by the current model and we compute the total cross-entropy loss for token $t_i$ as:
%%%%Another source of signal of informativeness can be derived from how badly the language model can predict the token. Therefore, we propose the other sampling strategy shown in Figure \ref{fig:example}. Firstly, we set an initial masking probability $p_i$ for each token $t_i \in T$ as $\forall i, p_i = 1$. Then, we employ training the language model by using the sampling probability in the dictionary. During each mini-batch, the masked tokens will be predicted by the current model and we calculate the total cross-entropy loss for token $t_i$ as:
\begin{equation}
\small
    %L_{t_i, \tilde{t_i}} = - \sum_{t_i \in T} logP(\Tilde{t_i} \mid x, \theta)
    % L_{t_i, \tilde{t_i}} = - \sum_{t_i \in T} logP(\Tilde{t_i} \mid x, \theta)
    L_{t_i} = - logP(t_i \mid x, \theta)
\end{equation}
\noindent where $x$ denotes the input masked sequence and $\theta$ denotes the parameters of the current masked language model. 
Then, we use $L_{t_i}$ to compute the sampling weight $wt(t_i)$ based on Eqn.~\ref{eq:dyn_weight}.
%where $x$ is the input masked sequence and $\theta$ is the parameters of the language model. $T$ denotes tokens in a dataset. We further utilize $L_{t_i, \tilde{t_i}}$ to get the sampling probability $p_i$ for token $t_i$ by applying the transformation:
\begin{equation}
\small
wt(t_i) = exp(\frac{L_{t_i}}{\tau})
\label{eq:dyn_weight}
\end{equation}
\noindent where $\tau$ is a temperature parameter with default as 0.2. Finally, for the next mini-batch, we compute the sampling probability $p(t_i)$ for token $t_i$ by normalizing $wt(t_i)$ following Eqn.~\ref{eq:freq_prob}.

%\begin{equation}
%\label{eq:dyn_prob}
%\small
%p(t_i) = \frac{wt(t_i)}{\sum_{j=1}^{n}{wt(t_j)}}
%\end{equation}

%$n$ is the number of tokens in a sentence.

The sampling weights computed by Eqn.~\ref{eq:dyn_weight} are larger for tokens with higher cross-entropy prediction loss, i.e., tokens that are poorly learned in the current masked language model and are often rare tokens; the sampling weights are smaller for tokens with lower cross-entropy loss, i.e., tokens that are relatively better learned.  We design the sampling weight function $wt(t_i)$ as Eqn.~\ref{eq:dyn_weight} to enlarge the variance of sampling weights between different tokens and to further boost up sampling probabilities of rare tokens.  During each iteration in pre-training, the weight dictionary is updated with the latest sampling weights $wt(t_i)$ for each token $t_i$. For the next iteration, for a sequence $s={t_1, t_2, ..., t_n}, s \in S$, the sampling probability of choosing to mask each token is computed using the updated weight dictionary by Eqn.~\ref{eq:freq_prob}.

\vspace{-4mm}
\section{Experiments}
\label{sec:experiment}
\vspace{-2mm}

\begin{table}[]
\centering
\scalebox{0.65}{
\begin{tabular}{@{}l|lllllll|l@{}}
\toprule[1.5pt]
Method                 & STS12 & STS13 & STS14 & STS15 & STS16 & STS-B & SICK-R & Avg.  \\ \midrule
BERT              & 39.70 & 59.38 & 49.67 & 66.03 & 66.19 & 53.87        & 62.06           & 56.70 \\
BERT-CP &41.00  &60.02  &51.11 &68.43 &64.59  &56.32   &62.07    &57.65 \\
WSBERT\_Freq         & 42.60 &61.32 &52.04 &69.84 &66.61 &59.89  &61.94 &59.18 \\
WSBERT\_Dynamic               & 47.80 & 67.28 & 57.13 & 71.41 & 68.87 & 65.28        & 64.90           & 63.24 \\ \midrule
BERT-Whitening      & 54.28 & {\textbf{78.07}} & {\textbf{65.44}} & 64.83 & 70.16 & 71.43        & 62.23           & 66.43 \\
WSBERT-Whitening   & 55.14 & {\ul\textbf{78.45}} & {\ul\textbf{66.13}} & 65.47 & {\textbf{70.68}} & {\textbf{71.98}}       & 61.91          & 67.10 \\ \midrule
BERT + Prompt$\dag$ &{\textbf{60.96}} &73.83 &62.18 &{\textbf{71.54}} &68.68 &70.60 &{\ul\textbf{67.16}} & {\textbf{67.85}} \\
WSBERT + Prompt & {\ul\textbf{63.03}} & 71.66 & 63.80 & {\ul\textbf{75.32}} & {\ul\textbf{76.67}} & {\ul\textbf{74.79}}        & {\textbf{65.32}}         & {\ul\textbf{70.08}} \\  \bottomrule[1.5pt]
\end{tabular}
}
\caption{\small{Sentence representation performance on STS tasks. The reported score is Spearman’s correlation coefficient between the predicted similarity and the gold standard similarity scores. The best results are both underlined and in bold. WSBERT without a subscript refers to WSBERT\_Dynamic. Performance of BERT-Whitening is from the model learned on the full target dataset with embedding size 256 (train+development+test)~\cite{whitening}. $\dag$ denotes the best manual-prompt results cited from \cite{promptbert}}.
%The performance of WSBERT on STS tasks. The best results are underlined and in bold. BERT-Whitening is reported with the one learned over the full target dataset. The reported score is Spearman’s correlation coefficient between the predicted similarity and gold standard similarity scores. $\dag$ records the best manual-prompt result from \cite{jiang2022promptbert}.
}
\label{table:sts}
\end{table}

\begin{table}[]
\centering
\scalebox{0.9}{%
\begin{tabular}{@{}c|c|c|c@{}}
\toprule[1.5pt]
Dataset      & BERT            & BERT-CP     & WSBERT                     \\ \midrule
MNLI         & $84.30_{\pm0.26}$                & ${84.26_{\pm{0.19}}}$       & {$\mathbf{84.42_{\pm{0.35}}}$}       \\
QQP          & $91.31_{\pm{0.04}}$                & $90.94_{\pm{0.59}}$       & {$\mathbf{91.43_{\pm{0.05}}}$}       \\
QNLI         & {$\mathbf{91.47_{\pm{0.01}}}$} & $91.32_{\pm{0.17}}$       & $91.14_{\pm{0.17}}$                      \\
SST-2        & {$\mathbf{92.86_{\pm{0.13}}}$} & $92.78_{\pm{0.43}}$       & $91.35_{\pm{0.47}}$                      \\
CoLa         & $56.47_{\pm{0.65}}$                & $57.44_{\pm{0.95}}$       & {$\mathbf{58.29_{\pm{0.33}}}$}       \\
STS-B        & $89.68_{\pm{0.26}}$                & $89.52_{\pm{0.37}}$       & {$\mathbf{89.86_{\pm{0.18}}}$}       \\
MRPC         & $86.13_{\pm{1.63}}$                & $85.13_{\pm{0.53}}$       & {$\mathbf{88.20_{\pm{2.39}}}$}       \\
RTE          & $69.23_{\pm{0.4}}$                & $67.25_{\pm{1.84}}$       & {$\mathbf{70.89}_{\pm{0.17}}$}       \\ \midrule
\textbf{AVG} & $82.68_{\pm0.33}$          & $82.33_{\pm0.32}$ &  $\mathbf{83.20}_{\pm0.10}$ \\ \bottomrule[1.5pt]
\end{tabular}
}
\caption{\small{GLUE Validation results from \emph{BERT-base-uncased} (BERT-base),  \emph{BERT-base-uncased} continually pre-trained (BERT-CP), and Weighted-Sampled BERT (WSBERT). BERT-CP and WSBERT both continually train on BERT with the same training settings. WSBERT refers to WSBERT\_Dynamic. The best results for each dataset and AVG are in bold.}}
\label{table:glue}
\end{table}

 % We conduct two sets of experiments. Firstly, we investigate the quality of unsupervised sentence representations using WSBERT, by evaluating on the commonly used STS tasks. Secondly, we investigate the efficacy of Weighted Sampling on the transfer learning capability of BERT, by fine-tuning and evaluating WSBERT on the GLUE benchmark. We also design ablation studies to analyze the impact of WSBERT on token embeddings for rare tokens and common tokens.
 We conduct two experiments: evaluating unsupervised sentence representations using WSBERT on STS tasks, and evaluating the efficacy of Weighted Sampling on BERT's transfer learning capability by fine-tuning WSBERT on the GLUE benchmark. Ablation studies are also designed to analyze the impact of WSBERT on token embeddings for rare and common tokens.

\subsection{Datasets and Implementation Details}
\label{sub:dataset}
For WSBERT, we continue pre-training on \verb!bert-base-uncased! (BERT)\footnote{\url{https://huggingface.co/bert-base-uncased}}, with Weighted Sampling on the WikiText dataset (+100M tokens) \footnote{\url{https://huggingface.co/datasets/wikitext}}. We set the learning rate $5 \times 10^{-5}$ and train 10 epochs. We use 4 NVIDIA V100 GPUs to train WSBERT with batch size 8 per device and gradient accumulation as 8. We use the WordPiece tokenizer as in ~\cite{bert} and token frequency is based on the tokenized Wikitext. STS tasks contain STS 2012-2016, STS benchmark, and SICK-Relatedness datasets. GLUE includes eight datasets~\cite{glue}.  To analyze the effect of continual pre-training without Weighted Sampling, we also continue pre-training on BERT with the same random sampling as for BERT on the same WikiText data and with the same pre-training setting as WSBERT. We denote the resulting model \textbf{BERT-CP}. We compare fine-tuning performance of \verb!bert-base-uncased! (BERT), BERT-CP, and WSBERT on GLUE. 
For each model on each GLUE task, we run three runs with different random seeds; for each run, we conduct a grid search on hyperparameters on GLUE validation set among ${2 \times 10^{-5},3 \times 10^{-5},5 \times 10^{-5}}$ learning rate and ${5, 10}$ epochs. The other hyperparameters are the same for the three models: we use 1 V100 with batch size 32 per device, warm-up ratio 0.06, and weight-decay 0.01. We then report the mean and standard deviation of the best results from three runs in Table~\ref{table:glue}.

\vspace{-3mm}
\subsection{Main Results}
%\subsection{Performance Analyses}
\label{sub:main-results}
\vspace{-1mm}
\noindent \textbf{Semantic Textual Similarity}
Table~\ref{table:sts} shows the main STS results. All models in the table are of BERT base size. We report results of WSBERT with \emph{Frequency Weighted Masking} and \emph{Dynamic Weighted Masking}, denoted WSBERT\_Freq and WSBERT\_Dynamic. The first group in Table~\ref{table:sts} shows that WSBERT\_Dynamic outperforms BERT and BERT-CP by \textbf{6.54} and \textbf{5.59} absolute, significantly improving the quality of sentence embeddings of PLMs. WSBERT\_Dynamic outperforms WSBERT\_Freq by 4.06 absolute, showing that Dynamic Weighted Sampling is more effective than sampling only based on token frequency. BERT\_Whitening~\cite{whitening}, as a calibration method, is compatible with WSBERT. The second group in Table~\ref{table:sts} shows that although WSBERT\_Dynamic yields a lower average score on STS compared to BERT\_Whitening, WSBERT\_Dynamic could be effectively combined with Whitening and further improve the performance of WSBERT\_Dynamic to \textbf{67.10}. 
We also investigate enhancing BERT and WSBERT with prompt. Different from previous works using a single \verb![MASK]! in prompts~\cite{promptbert}, we transform a sentence using manual prompt templates with multiple \verb![MASK]!\footnote{The best prompt is designed as The sentence: ${[}X{]}$ means ${[}MASK{]}$ and also means ${[}MASK{]}$.}.  
Furthermore, instead of extracting and averaging representations of the masked tokens as final sentence embeddings~\cite{promptbert}, we encode the whole transformed sentence and compute sentence embedding by average pooling all token embeddings of the sentence.
%Furthermore, instead of extracting and averaging representations of the masked tokens as in \cite{jiang2022promptbert}, we encode the whole transformed sentence into a representation by average pooling. 
The third group in Table~\ref{table:sts} shows that prompt-enhanced WSBERT achieves \textbf{70.08}. These results demonstrate that Weighted Sampling improves sentence representations generated by PLMs and combining WSBERT with Whitening and prompts further improves sentence embeddings.  We did not compare WSBERT to the SOTA models on STS, i.e., sentence-level contrastive learning (CL) based models such as SimCSE \cite{simcse}, since prior works~\cite{tacl} and our studies show sentence-level CL based models hurt transfer learning capability. We observe absolute 0.5 degradation on GLUE AVG score from SimCSE-BERT compared to BERT. However, as shown in the following GLUE experiments, WSBERT both enhances sentence representations of BERT and improves transfer learning capability.

\noindent \textbf{GLUE Evaluation}
As shown in Table~\ref{table:glue}, WSBERT achieves the best average GLUE score compared to BERT and BERT-CP, outperforming BERT by {\textbf 0.52} absolute. 
%Although gains are primarily from smaller datasets such as RTE, CoLA, and STS-B, 
WSBERT maintains competitive performance on MNLI and QQP and outperforms BERT on all other tasks. In contrast to models such as SimCSE, Dynamic Weighted Sampling improves the transfer learning capability while enhancing sentence representations. Compared to BERT, BERT-CP degrades GLUE AVG by 0.35 absolute while WSBERT outperforms BERT-CP by 0.87 absolute. These results prove that the gain of WSBERT over BERT is from continual pre-training with Dynamic Weighted Sampling instead of just continuing pre-training BERT using random sampling with more steps on the same WikiText dataset. BERT-CP degrades GLUE performance compared to BERT, probably because WikiText (373.28M data size) used for continual pre-training is much smaller and less diverse than the standard BERT pre-training dataset (Wikipedia and Bookscorpus, 16GB data size), which could hurt generalizability of PLMs.

%We conduct experiments to compare the performances of three BERT models with an evaluation of GLUE. We run with three different seeds to ensure the improvement is not from randomness and record the average score of three seeds in Table \ref{table:glue}. WSBERT achieves the best average score compared with the other two models, which outperforms BERT by an absolute 0.52, and also exceeds BERT-CP with 0.82. Therefore, continual pre-training BERT with a weighted-sampling strategy could enhance the baseline model with a better capability of transfer learning. Although gains are primarily from small datasets for instance RTE, CoLA, and STS-B, WSBERT still obtains competitive scores on MNLI and QQP. The comparison of BERT-CP and WSBERT proves that the gain from WSBERT over BERT on GLUE is not from continual pre-training of BERT with more steps on WikiText. BERT-CP degrades GLUE performance, probably because continue pre-training on WikiText (373.28M), much smaller than the standard pre-training data (Wikipedia and Bookscorpus, 16GB), hurts the generalizability of the pre-trained language model. WSBERT outperforms BERT-CP by +0.5 absolute on GLUE.
%Without pertaining, only fine-tuning the baseline model with a weighted-sampling strategy could bring a boost for the baseline model, thus saving computing resources.  

% \usepackage{booktabs}
% \usepackage[normalem]{ulem}

\vspace{-3mm}
\subsection{Analysis}
\vspace{-1mm}
As observed in~\cite{on_the_sentence, consert}, token embeddings of MLM-pretrained PLMs can be biased by token frequency, causing embeddings of high-frequency tokens to concentrate densely and low-frequency tokens to disperse sparsely. Inspired by these works, to analyze whether Weighted Sampling could indeed alleviate the frequency bias problem, we propose two approaches to analyze distributions of BERT, BERT-CP, and WSBERT in the representation space. We also discuss the training time for training MLM with and without weighted sampling.

%As mentioned in \cite{li2020sentence, consert}, token embeddings can be biased to token frequency, causing high-frequency words to concentrate densely and low-frequency words to disperse sparsely. In order to discover whether the weighted-sampling strategy could deal with the frequency bias issue, we propose two approaches to analyze the tokens distributions of BERT, BERT-CP, and WSBERT in embedding space.

\begin{table}[]
\centering
\begin{tabular}{@{}c|c|c@{}}
\toprule
Method            & Rare Tokens           & Common Tokens         \\ \midrule
BERT & 14.97                & 64.82                \\
BERT-CP          & 14.86                & 64.95                \\
WSBERT          & {\ul \textbf{15.60}} & {\ul \textbf{65.54}} \\ \bottomrule
\end{tabular}
\caption{\small{Portion of \emph{common tokens} (high-frequency tokens) in the nearest neighbors of \emph{rare tokens} (low-frequency tokens) and \emph{common tokens}. We first sort tokens in the WikiText vocabulary by frequency in descending order. Then we select the tokens with ranks ranging from 10K-20K as rare tokens while choosing the Top-10K tokens as common tokens. We choose the 10 nearest neighbors decided by the Euclidean distance between representations of the target token and other tokens.}}
\label{table:nn}
\end{table}

\noindent \textbf{Nearest Neighbors}
We investigate the portion of common tokens in the nearest neighbors (NN) of rare tokens, denoted $P_{rare}$, and the portion of common tokens in NN of common tokens, denoted $P_{common}$ (rare and common tokens are defined in Table~\ref{table:nn} caption). The larger portion of common tokens in NN of rare/common tokens indicates more concentrated token distributions and hence smaller frequency bias in the token embeddings. In Table~\ref{table:nn}, $P_{rare}$/$P_{common}$ for WSBERT increase by 0.63/0.72 over those of BERT and 0.74/0.59 over those of BERT-CP,  suggesting that WSBERT has more concentrated token distributions and smaller frequency bias in token embeddings compared to BERT and BERT-CP,  and common tokens are also more concentrated in WSBERT than BERT and BERT-CP. 

%The portion of common tokens of WSBERT increases by 0.74 and 0.59 compared with that of BERT-base and BERT-CP. The improvement is significant in the neighbors of rare tokens, proving the distribution of rare tokens embeddings in WSBERT are more concentrated and the embeddings of rare tokens contain more semantics. Thus the frequency bias is alleviated by the weighted-sampling strategy to some extent.

\begin{table}[]
\centering
\scalebox{0.7}{
\begin{tabular}{@{}cc|cccc@{}}
\toprule[1.5pt]
\multicolumn{2}{c|}{\textbf{Rank of token frequency}}                              & 0-100  & 100-500 & 500-5k & 5k-10k \\ \midrule
\multicolumn{1}{c|}{\multirow{3}{*}{Mean $\ell2$-norm}}            & BERT & 0.9655 & 1.0462  & 1.2150 & 1.3639  \\
\multicolumn{1}{c|}{}                                         & BERT-CP            & 0.9597 & 1.0428  & 1.2141 & 1.3647  \\
\multicolumn{1}{c|}{}                                         & WSBERT            & \textbf{0.9562} & \textbf{1.0385}  & \textbf{1.2112} & \textbf{1.3621}  \\ \midrule
\multicolumn{1}{c|}{\multirow{3}{*}{Mean k-NN $\ell2$-norm (k=3)}} & BERT & 0.6972 & 0.7782  & 0.8188 & 0.8953  \\
\multicolumn{1}{c|}{}                                         & BERT-CP            & 0.6913 & 0.7750  & 0.8180 & 0.8963  \\
\multicolumn{1}{c|}{}                                         & WSBERT            & \textbf{0.6883} & \textbf{0.7724}  & \textbf{0.8154} & \textbf{0.8929}  \\ \midrule
\multicolumn{1}{c|}{\multirow{3}{*}{Mean k-NN $\ell2$-norm (k=5)}} & BERT & 0.8007 & 0.8868  & 0.9327 & 1.0083  \\
\multicolumn{1}{c|}{}                                         & BERT-CP            & 0.7936 & 0.8833  & 0.9319 & 1.0096  \\
\multicolumn{1}{c|}{}                                         & WSBERT            & \textbf{0.7899} & \textbf{0.8800}  & \textbf{0.9287} & \textbf{1.0056}  \\ \midrule
\multicolumn{1}{c|}{\multirow{3}{*}{Mean k-NN $\ell2$-norm (k=7)}} & BERT & 0.8590 & 0.9458  & 0.9932 & 1.0671  \\
\multicolumn{1}{c|}{}                                         & BERT-CP            & 0.8513 & 0.9422  & 0.9924 & 1.0685  \\
\multicolumn{1}{c|}{}                                         & WSBERT            & \textbf{0.8471} & \textbf{0.9386}  & \textbf{0.9888} & \textbf{1.0642}  \\ \bottomrule[1.5pt]
\end{tabular}
}
\caption{\small{The mean $\ell2$-norm calculated for each bin of tokens with ranking ranges based on token frequency in WikiText. 
%We partition tokens according to their ranking ranges based on token frequency in WikiText into bins. 
Common tokens occupy a higher ranking while rare tokens are in low rankings. 
%segment
A lower mean $\ell2$-norm suggests that the token embeddings in that bin are more concentrated. }}
\label{table:l2}
\end{table}

\noindent \textbf{Token Distribution}
Inspired by~\cite{on_the_sentence}, we compute the mean $\ell2$-norm between token embeddings and the origin for the three models to analyze token distributions. As shown in the first row of Table~\ref{table:l2}, although common tokens are close to the origin and rare tokens are distributed far away from the origin, the smaller mean $\ell2$-norm indicates the token embeddings of WSBERT are more concentrated than BERT and BERT-CP. The $\ell2$-norms of WSBERT are smaller on all the bins than those of BERT,  suggesting that both common tokens and rare tokens are closer to the origin in WSBERT than BERT.
Furthermore, WSBERT tokens in each bin are more compact than BERT and BERT-CP as shown by the smaller mean $k$-NN $\ell2$-norm for each $K$ in Table~\ref{table:l2}, which indicates that the embedding space of WSBERT is less sparse than BERT and BERT-CP. Sparsity in the embedding space may cause poorly defined semantic meanings~\cite{on_the_sentence}, hence the gains in sentence embeddings and transfer learning capability from WSBERT over BERT may also be attributed to sparsity reduction by WSBERT in the embedding space.

\noindent \textbf{Training Time}
Calculating the sampling weight for each token during training takes extra time. Training a MLM without weighted sampling takes 11 hours while training with weighted sampling takes 20 hours. The extra time can be reduced through optimizations such as implementing parallel writing to the dictionary and using an in-memory vector to accelerate the reading and writing processes.
% The training time of MLM without weighted sampling for 10 epochs is 11 hours while MLM with weighted sampling takes 20 hours. However, this trade-off between efficiency and performance is acceptable.
%Inspired by~\cite{li2020sentence}, we compute the mean $\ell2$ distance between token embeddings and the origin for three models to analyze their token distribution. Although common tokens are still close to the origin and rare tokens are distributed far away from the origin, the mean $\ell2$ norm indicates the token embeddings of WSBERT are more concentrated compared to BERT and BERT-CP. Because, the $\ell2$ norm of WSBERT is smaller on all of the segments compared with BERT, which also represents that both common tokens and rare tokens are closer to the origin.  Besides, WSBERT tokens in each segment are more compacted than BERT and BERT-CP with regard to the mean $k$-NN $\ell2$-norm.  It means that the embedding space of WSBERT is not as sparse as BERT and BERT-CP. Sparsity will cause some "holes" in the embedding space which could make the semantic meaning poorly defined \cite{li2020sentence}. In this way, WSBERT continually trained with weighted-sampling strategy could make BERT embedding space with a low sparsity, which advances BERT in GLUE and STS task.
% As for the mean k-NN $\ell2$-norm, WSBERT has a denser token distribution than BERT and BERT-CP in all of the segments.

%\subsubsection{Single Value}
\vspace{-4mm}
\section{Conclusion}
\vspace{-2mm}
We propose two Weighted Sampling methods to alleviate the frequency bias issue in Masked Language Modeling for pre-training language models. Extensive experiments show Weighted Sampling improves both sentence representations and the transferability of pre-trained models. We also analyze the token embeddings to explain how Weighted Sampling works. Future work includes investigating other dynamic sampling methods and exploring training objectives with a penalty for frequency bias.

\footnotesize
\bibliographystyle{IEEEbib}
\bibliography{refs}

\end{document}